\documentclass{article}

\usepackage[preprint]{neurips_2025}

\usepackage[utf8]{inputenc} 
\usepackage[T1]{fontenc}    
\usepackage{hyperref}       
\usepackage{url}            
\usepackage{booktabs}       
\usepackage{amsfonts}       
\usepackage{nicefrac}       
\usepackage{microtype}      
\usepackage{xcolor}         
\usepackage{mymacro}
\usepackage{algorithm,algorithmic}

\title{$\fedada$: Doubly Adaptive Federated Learning}

\author{%
  Shokichi Takakura\\
  LY Corporation\\
  \texttt{stakakur@lycorp.co.jp} \\
  \And
  Seng Pei Liew \\
  LY Corporation \\
  \texttt{sliew@lycorp.co.jp} \\
  \And
  Satoshi Hasegawa \\
  LY Corporation \\
  \texttt{satoshi.hasegawa@lycorp.co.jp} \\
}

\begin{document}

\maketitle

\begin{abstract}
    Federated learning is a distributed learning framework where clients collaboratively train a global model without sharing their raw data.
    FedAvg is a popular algorithm for federated learning, but it often suffers from slow convergence
    due to the heterogeneity of local datasets and anisotropy in the parameter space.
    In this work, we formalize the central server optimization procedure through the lens of mirror descent and propose a novel framework,
    called \textit{$\fedada$}, which adaptively selects the global learning rate based on both inter-client and coordinate-wise heterogeneity in the local updates.
    We prove that our proposed doubly adaptive step-size rule is minimax optimal and provide a convergence analysis for convex objectives.
    Although the proposed method does not require additional communication or computational cost on clients,
    extensive numerical experiments show that our proposed framework outperforms baselines in various settings and is robust to the choice of hyperparameters.
\end{abstract}

\section{Introduction}
Federated Learning (FL)~\citep{konevcny2016federated} is a distributed optimization framework where multiple clients
collaboratively train a global model under the coordination of a central server.
In FL, clients have their own local datasets and only send model updates to a central server
and never share their raw data, which enhances the privacy of the data.
In FL, there are two primary categories: \textit{cross-silo} FL and \textit{cross-device} FL~\citep{kairouz2021advances}.
In this paper, we focus on cross-device FL, which is more challenging due to the limited computational resources and communication bandwidth on clients.

Federated Averaging (FedAvg)~\citep{mcmahan2017communication} is one of the most popular algorithms for FL
due to its simplicity, stateless properties, and communication efficiency.
In FedAvg, clients perform multiple local training steps before sending the local updates to the server,
which significantly reduces the communication cost to train a global model.
However, FedAvg often suffers from slow convergence due to (1) the \textit{client heterogeneity} and (2) \textit{gradient heterogeneity}.
The former refers to non-i.i.d. data distribution across clients,
which leads to so-called \textit{client drift error}~\citep{kairouz2021advances}.
The latter refers to the anisotropic nature of gradients,
meaning that gradients have different scales or importance across different parameter dimensions,
which often hinders the convergence of SGD~\citep{zhang2020adaptive,zhang2024transformers,tomihari2025understanding}.

A line of work has dealt with the client heterogeneity by introducing control variates
to reduce the client drift~\citep{karimireddy2020mime,karimireddy2019error,mishchenko2022proxskip}.
While these methods are effective in \textit{cross-silo} FL,
it is not practical in \textit{cross-device} FL, where clients have limited computational resources and communication bandwidth
since they require clients to be stateful and increase the communication and computational cost on clients.
Recently, \citet{jhunjhunwala2023fedexp} have proposed FedExP, which accelerates FedAvg by selecting the global learning rate adaptively to the client heterogeneity.

To deal with the gradient heterogeneity, \citet{reddi2021adaptive} have proposed a federated version of adaptive optimizers
including FedAdagrad, FedAdam, FedYogi
inspired by the success of adaptive methods in centralized optimization.
These methods adaptively adjust the coordinate-wise learning rate based on the historical local updates.
They have shown the coordinate-wise adaptivity can improve the performance of FL especially for tasks with sparse gradients.

Although adaptivity to both client and gradient heterogeneity is shown to be crucial to achieve fast convergence~\citep{jhunjhunwala2023fedexp,reddi2021adaptive},
most of existing works focus on either client or gradient heterogeneity,
and it is not straightforward to combine them since they have been proposed from distinct conceptual standpoints.
Consequently, it is still unclear how to incorporate two types of adaptivity
without additional computational and communication cost.
Thus, we pose the following question:
\textit{How can we unify the adaptivity to the client and gradient heterogeneity, and design a doubly adaptive global update procedure?}

To answer this question, we formulate the global update procedure through the lens of mirror descent,
which is a generalization of gradient descent and offers a novel perspective for designing global update procedure.
From this perspective, we unify existing adaptive methods and propose a novel framework, called \textit{$\fedada$}, which adaptively selects the global learning rate based on both the client and gradient heterogeneity.
We numerically and theoretically show that dual adaptivity is essential to achieve better performance.
In contrast to some existing methods~\citep{karimireddy2020scaffold,qu2022generalized}, $\fedada$ does not require additional computational or communication cost on clients.
See Table~\ref{tab:comparison} for the comparison with existing adaptive FL algorithms.
We would like to emphasize that $\fedada$ is orthogonal to the existing methods which improve the performance by modifying the local training procedure.
Thus, $\fedada$ can be combined with them to further improve the performance.

\paragraph{Main contributions}
Our contribution can be summarized as follows:
\begin{itemize}
    \item We formulate the global update procedure from the mirror descent perspective and propose a novel framework, called $\fedada$,
          which adaptively selects the global learning rate based on both the inter-client and coordinate-wise heterogeneity in local updates.
    \item We show that the update rule of $\fedada$ is minimax optimal under the approximate projection condition and provide the convergence analysis.
    \item We conduct extensive experiments on various datasets and show that FedDuA consistently outperforms existing adaptive methods. We also show that $\fedada$ is robust to the choice of hyperparameters due to its dual adaptivity.
\end{itemize}

\begin{table}[t]
    \centering
    \caption{Comparison of adaptive methods in FL.}
    \begin{tabular}{lccc}
        \toprule
                                                 & \multicolumn{2}{c}{Adaptivity} &                                     \\
        Algorithms                               & Coordinate-wise                & Inter-client & No extra client cost \\
        \midrule
        AdaAlter~\citep{xie2019local}            & \checkmark                     &              &                      \\
        SCAFFOLD~\citep{karimireddy2020scaffold} &                                & \checkmark   &                      \\
        FedOpt~\citep{reddi2021adaptive}         & \checkmark                     &              & \checkmark           \\
        FedExP~\citep{jhunjhunwala2023fedexp}    &                                & \checkmark   & \checkmark           \\
        \fedada~(ours)                           & \checkmark                     & \checkmark   & \checkmark           \\
        \bottomrule
    \end{tabular}
    \label{tab:comparison}
\end{table}
\subsection{Other Related Work}
\paragraph{Adaptive Methods}
Adaptive methods like AdaGrad~\citep{duchi2011adaptive}, Adam~\citep{kingma2015adam}, and Yogi~\citep{zaheer2018adaptive} have been widely used in centralized optimization.
Inspired by the success in centralized optimization, several works have utilized adaptive methods in FL.
For instance, \citet{xie2019local} have proposed AdaAlter, which replaces the local SGD with an adaptive optimizer such as AdaGrad.
On the other hand, \citet{reddi2021adaptive} have proposed to use such adaptive methods as a global optimizer in FL.
Several works~\citep{lee2024efficient,wang2021local} have unified the above strategies and used adaptive optimizers at both the client and server.
However, these methods are not adaptive to the heterogeneity among clients.

\paragraph{Mirror Descent}
Mirror descent is a generalization of gradient descent~\citep{hazan2016introduction} and has been adopted in various applications including online learning~\citep{hazan2016introduction}, reinforcement learning~\citep{tomar2022mirror}, and differentially private optimization~\citep{odeyomi2021privacy, amid2022public}.
In the context of FL, \citet{yuan2021federated} have proposed FedMid and FedDualAvg with local mirror descent,
but these methods are tailored for composite optimization problems and do not consider adaptive step-size selection, which is the focus of our work.

\paragraph{Notation}
For a vector $x \in \R^d$, $[x]_k$ denotes the $k$-th element of $x$, $\norm{x}_p$ denotes the $p$-norm of $x$,
and $\norm{x}_{G}$ denotes $\sqrt{x^\top G x}$ for a positive semi-definite matrix $G \in \R^{d \times d}$.
We use $\norm{x}$ as a shorthand for $\norm{x}_2$.
For $s \in \R^d$, we write $\sqrt{s}$, $s^{-1}$ and $s + a~(a\in \R)$ as the element-wise square root, inverse and addition respectively.

\section{Problem Formulations and Preliminaries}
In this paper, we consider the following federated learning problem:
\begin{align*}
    \min_{w \in \R^d} F(w) := \frac{1}{M}\sum_{i=1}^M F_i(w),
\end{align*}
where $F_i(w) = \Expec[z \sim \mathcal{D}_i]{f_i(w, z)}$ is the loss function and $\mathcal{D}_i$ is the data distribution of the $i$-th client.
The number of clients is denoted by $M$ and the dimension of the parameter is denoted by $d$.

\paragraph{FedAvg} To solve the above optimization problem, we consider FedAvg~\citep{mcmahan2017communication}, which is a standard algorithm for FL.
At each round $t$, the server sends the global model $w_t$ to all clients.
Then, each client performs $\tau$-steps of local training using SGD to obtain the local updates $\{\Delta^t_i\}_{i=1}^M$ as follows:
\begin{align*}
    \text{Run Local SGD: }          & w^{t}_{i, k + 1} = w^t_{i, k} - \eta_l \nabla f_i(w^t_{i, k}, z_k)~(k=0, \dots, \tau - 1) \\
    \text{Calculate Local Update: } & \Delta_i^t                 = w^t_{i, \tau} - w_t,
\end{align*}
where $w^t_{i, 0} = w_t$ and $z_k \sim \mathcal{D}_i$.
When clients complete the local training, they send the local updates to the server
and the server aggregates the local updates to obtain the global update $\bar \Delta_t = \frac{1}{M}\sum_{i=1}^M \Delta_i^t$.
In FedAvg, the central server updates the global model as follows:
\begin{align*}
    \text{Global Update (FedAvg): } & w_{t+1} = w_t + \eta_g\bar \Delta_t,
\end{align*}
where $\eta_g$ is the global learning rate. Vanilla FedAvg uses $\eta_g = 1$, but in practice, $\eta_g > 1$ is often used to improve the convergence.
FedAvg enjoys some favorable properties such as statelessness and communication efficiency,
but it often suffers from slow convergence due to 1) the anisotropic nature of the local updates and 2) the heterogeneity of client data distributions.
We call the former \textit{gradient heterogeneity} and the latter \textit{client heterogeneity}.
Considering limited computational resources and communication bandwidth on clients, it is desirable to design an algorithm on a central server to overcome the above issues
without changing the local training procedure.

\paragraph{FedOpt}
To deal with the gradient heterogeneity, \citet{reddi2021adaptive} have introduced a general framework called FedOpt.
In this framework, the aggregated update $\bar \Delta_t$ is regarded as a pseudo-gradient and adaptive methods such as AdaGrad and Adam are used as a global optimizer.
General update rule of FedOpt is given by
\begin{align*}
    \text{Global Update (FedOpt): } & w_{t+1} = w_t + \eta_g G_t^{-1}v_t
\end{align*}
where $G_t:= \diag(\sqrt{s_t} + \epsilon)~(s_t \in \R^d, \epsilon > 0)$ is a time-dependent preconditioner and $v_t$ is the aggregated update or momentum term.
Here, $\epsilon > 0$ is a small constant for numerical stability.
In FedAdagrad and FedAdam, $s_t$ and $v_t$ are defined as
\begin{align*}
    \text{FedAdagrad: } & s_t = s_{t-1} + \bar \Delta_t^2,~v_t = \bar \Delta_t.                                                    \\
    \text{FedAdam: }    & s_t = \beta_2 s_{t-1} + (1-\beta_2)\bar \Delta_t^2,~v_t = \beta_1 v_{t-1} + (1 - \beta_1) \bar \Delta_t,
\end{align*}
where $s_{-1}, v_{-1} = 0$ and $\beta_1, \beta_2 \in [0, 1)$ are hyperparameters.

\paragraph{FedExP}
To deal with the client heterogeneity, \citet{jhunjhunwala2023fedexp} have proposed FedExP, which adaptively selects the global learning rate $\eta_g$ in FedAvg based on the client heterogeneity.
FedExP updates the global model as follows:
\begin{align*}
    \text{Global Update (FedExP): } & w_{t+1} = w_{t} + \eta_g^t \bar \Delta_t
\end{align*}
with adaptive global learning rate $\eta_g^t = \frac{\frac{1}{2M}\sum_{i=1}^M \norm{\Delta_i^t}^2}{\norm{\bar \Delta_t}^2 + \epsilon_g}$.
Here, $\epsilon_g$ is a small constant to avoid blow-up of the step size.
If clients perform one step of local training with full-batch SGD and $\epsilon_g = 0$, $\Delta_i^t = \eta_l\grad F_i(w_t)$ and $\eta_g^t$ is reduced to $\frac{\frac{1}{2M}\sum_{i=1}^M \norm{\grad F_i(w_t)}^2}{\norm{\grad F(w_t)}^2}$,
which is known as the measure of the heterogeneity among clients~\citep{haddadpour2019convergence}.
Note that the above formula is tailored for FedAvg and not applicable to general FedOpt algorithms.
This is because the learning rate in FedExP is derived based on norms of raw local and global updates,
whereas FedOpt transforms these updates before applying them, making the original FedExP formulation incompatible.

\section{Generalized Formulation and Proposed Method}
Previous works have considered adaptivity to 1) gradient heterogeneity and 2) client heterogeneity separately.
This paper unifies these two types of adaptivity through a generalized formulation of the global update procedure.

\paragraph{Mirror descent formulation}
As discussed in the original paper of Adagrad~\citep{duchi2011adaptive}, adaptive methods
can be viewed as a generalized version of Mirror Descent.
That is, the update rule of FedOpt can be written as
\begin{align*}
    \begin{split}
        \text{Mirror Map: }           & \theta_t         = \grad \psi_t(w_t),      \\
        \text{Dual Variable Update: } & \theta_{t+1}  = \theta_t + \eta_g v_t,     \\
        \text{Inverse Mirror Map: }   & w_{t+1}  = \grad \psi_t^{-1}(\theta_{t+1})
    \end{split}
\end{align*}
where $\psi_t(x) = \frac{1}{2}x^\top G_tx $ is a time-dependent distance-generating function
and $\grad \psi_t(x)$ is called the mirror map.
We define $\theta_t := \grad \psi_t(w_t)$ as the dual variable and $\phi_t(\theta) := \max_w \lrangle{w, \theta}-\psi_t(w)$ as the convex conjugate of $\psi_t$.
Note that the distance-generating function for adaptive methods is often chosen as the quadratic form
but general smooth and strictly convex functions can be used in this formulation.
In this paper, we focus on the quadratic case but also consider the general convex case
and our proposed framework can be applied to them. For simplicity, we assume that $\psi_t$ and $\phi_t$ are defined on $\R^d$, smooth and strongly convex,
which is satisfied by the quadratic case with $G_t \succ 0$.

Given a smooth and strictly convex function $\psi_t(x)$,
geometry of the parameter space is naturally induced by the Bregman divergence
defined as
\begin{align*}
    D_{\psi_t}(x\mid y) = \psi_t(x) - \psi_t(y) - \grad \psi_t(y)^\top (x - y).
\end{align*}
The Bregman divergence can be viewed as a generalization of the Euclidean (squared) distance.
In fact, when $\psi_t(x) = \frac{1}{2}x^\top x$, the Bregman divergence reduces to the Euclidean distance.

\paragraph{Proposed Method}
A natural question is then how to choose the global learning rate $\eta_g$ to minimize the distance between $w_{t+1}$ and an optimal solution $w^*$.
In~\citet{jhunjhunwala2023fedexp}, the distance is measured by Euclidean distance in the parameter space.
However, this choice of distance is not always appropriate especially in modern machine learning tasks
since the geometry of the parameter space is often anisotropic~\citep{zhang2024transformers,tomihari2025understanding}.
Thanks to our mirror-descent formulation, we can use the Bregman divergence to measure the distance between two points.
That is, we consider the following minimization problem over $\eta_g$:
\begin{align*}
    \min_{\eta_g} D_{\psi_t}(w^* \mid w^{t+1}) = \min_{\eta_g} D_{\phi_t}(\theta_t + \eta_g v_t \mid \theta^*).
\end{align*}
Here, the equality follows from the duality~\citep{nielsen2007bregman}.

As discussed in~\citet{jhunjhunwala2023fedexp}, FedAvg can be interpreted as a generalized Projection onto Convex Sets (POCS) algorithm in overparameterized convex optimization problems,
where the set of minimizers for each local objective $S_i$ is convex and the objective is to find a common minimizer $w^* \in \cap_{i=1}^M S_i$.
That is, clients perform approximate projection onto $S_i$ through local SGD and the server aggragate them to find a shared minimizer $w^*$.
Inspired by the above relation, we consider the \textit{(strong) approximate projection condition} (A.P.C.):
\begin{assumption}\label{assumption:approximate-projection-condition}
    Let $w^* = \arg\min F(w)$ be the optimal solution of the global problem.
    \begin{align}
        \text{A.P.C.}        & \quad\frac{1}{M} \sum_{i=1}^M \norm{w_t + \Delta_i^t - w^*}^2 \leq \norm{w_{t} - w^*}^2, \label{eq:approximate-projection-condition}              \\
        \text{strong A.P.C.} & \quad\langle \bar \Delta_t, w^* - w_t\rangle \geq \frac{1}{M} \sum_{i=1}^M \norm{\Delta_i^t}^2 \label{eq:strong-approximate-projection-condition}
    \end{align}
\end{assumption}
Intuitively, A.P.C. means that the local models $w_t + \Delta_i^t$ after local training are closer to the optimal solution $w^*$ on average.
A.P.C. can be reduced to $\langle \bar \Delta_t, w^* - w_t\rangle \geq \frac{1}{2M} \sum_{i=1}^M \norm{\Delta^t_i}^2$.
Thus, strong A.P.C. is indeed a stronger condition than A.P.C.
As shown in Lemma 1 and Section C.4.1 of~\citet{jhunjhunwala2023fedexp}, A.P.C. is satisfied in the case of overparameterized convex optimization case
and strong A.P.C. is satisfied if the local training is the exact projection onto the set of optimal solutions of the local objective.
Note that we consider the above condition to motivate our proposed method and we prove later the convergence guarantee under much milder conditions.

Although local training at clients and approximate projection condition are agnostic to the choice of $\psi_t$,
we can derive the non-trivial lower bound on the optimal step size for general choice of $\psi_t$ for both cases with and without momentum.
\begin{theorem}[Lower Bound on the Optimal Step Size]\label{theorem:lower-bound}
    Let $v_t = \bar \Delta_t$, $h_t(\eta) := \odv{}{\eta} \phi_t(\theta_t + \eta v_t) - \langle v_t, w_t\rangle$,
    and $m_t = \frac{1}{2M}\sum_{i=1}^M \|\Delta_i^{t}\|^2$.
    Assume that A.P.C. in Assumption~\ref{assumption:approximate-projection-condition} holds and $v_t \neq 0$.
    Then, $\eta_g^* := \arg\min D_{\phi_t}(\theta_t + \eta_g \bar \Delta_t \mid \theta^*)$ is uniquely defined and satisfies
    \begin{align*}
        \eta_g^* \geq h_t^{-1}\left(m_t \right).
    \end{align*}
\end{theorem}
\begin{theorem}[Lower Bound on the Optimal Step Size with Momentum]\label{theorem:lower-bound-momentum}
    For $s=0, \dots, t$, let $v_s = (1-\beta_1)\bar \Delta_s + \beta_1 v_{s-1}~(v_{-1}=0)$,
    $h_s(\eta) := \odv{}{\eta} \phi_s(\theta_s + \eta v_s) - \langle v_s, w_s \rangle$,
    and $m_s = \frac{1-\beta_1}{2M}\sum_{i=1} ^M\|\Delta_i^s\|^2 + \frac{\beta_1}{2}m_{s-1}~(m_{-1} = 0)$.
    Assume that strong A.P.C. in Assumption~\ref{assumption:approximate-projection-condition} holds and $v_t \neq 0$.
    We further assume that at round $s=0, \dots, t-1$, $w_s$ is updated as Eq.~\eqref{eq:fedada-update} with $\eta_g^s \leq h_s^{-1}(m_s)$.
    Then, $\eta_g^* := \arg\min D_{\phi_t}(\theta_t + \eta_g v_t \mid \theta^*)$ is uniquely defined and satisfies
    \begin{align*}
        \eta_g^* \geq h_t^{-1}\left(m_t \right).
    \end{align*}
\end{theorem}
See Appendix~\ref{appendix:proof-lower-bound} and~\ref{appendix:proof-lower-bound-momentum} for the proof.
Here, $h_t^{-1}$ is well-defined since $\phi_t$ is assumed to be strongly convex and $v_t \neq 0$.
Similar analysis can be found in Section 3.2 and Lemma 11 of~\citet{jhunjhunwala2023fedexp}
but they only consider FedAvg(M) and $l^2$-norm. Our contribution is to extend this analysis to more general mirror descent framework.
To deal with the non-linearity of the global update in the parameter space,
we use the dual form of the Bregman divergence to derive the lower bound.
Note that it is essential to measure the distance with the Bregman divergence
since we cannot derive a non-trivial lower bound on the optimal step size if we use the Euclidean distance, which is \textit{not} compatible with the global update procedure.
The mirror descent formulation allows us to use an appropriate distance measure for a given global optimizer.

Motivated by the above result, we propose \textit{$\fedada$}, which uses the lower bound on the optimal step size as the global learning rate.
\begin{align}
    \begin{split}
        \text{Mirror Map: }         & \theta_t         = \grad \psi_t(w_t),      \\
        \text{$\fedada$ Update: }   & \theta_{t+1} = \theta_t + \eta_g^t v_t,    \\
        \text{Inverse Mirror Map: } & w_{t+1}  = \grad \psi_t^{-1}(\theta_{t+1})
    \end{split}\label{eq:fedada-update}
\end{align}
where $\eta_g^t := h_t^{-1}(m_t)$.
For quadratic case $\psi_t(x) = \frac{1}{2} x^\top G_t x$, the lower bound is calculated as
$
    \frac{m_t}{\norm{v_t}^2_{G_t^{-1}}}
$
in both cases, with and without momentum.
In practice, we can use $\frac{m_t}{\norm{v_t}^2_{G_t^{-1}} + \epsilon_g}$ with small constant $\epsilon_g > 0$ for stability.
We provide the detailed algorithm in Algorithm~\ref{alg:fedada2}.
As shown in Algorithm~\ref{alg:fedada2}, $\fedada$ is compatible with partial participation of clients, where only a subset of clients participate in each round.
For simplicity, we only provide the algorithms with Adagrad and Adam as the global optimizer, which we call FedDuAdagrad and FedDuAdam respectively.
Note that $\fedada$ framework is very versatile, allowing for the creation of new algorithms by combining it with various gobal optimizers through modification of $\psi_t$.
While we use SGD as a local optimizer here, we can use other local optimizers such as SCAFFOLD and Adam for local training.

\begin{algorithm}[t]
    \caption{$\fedada$}
    \label{alg:fedada2}
    \begin{algorithmic}[1]
        \REQUIRE Initial model $w_0$, local learning rate $\eta_l$, number of local steps $\tau$, small constants $\epsilon, \epsilon_g$
        \STATE Initialize $s_{-1} = 0 \in \R^d$, $v_{-1} = 0 \in \R^d, m_{-1}=0 \in \R$
        \FOR{$t = 0, 1, \dots, T-1$}
        \STATE Server sends $w_t$ to all clients
        \FOR{each client $i$ in parallel}
        \STATE Perform $\tau$ steps of local SGD to compute $\Delta_i^t$
        \STATE Send $\Delta_i^t$ to the server
        \ENDFOR
        \STATE Set $S_t$ as the set of participating clients at round $t$
        \STATE Server aggregates updates: $\bar{\Delta}_t = \frac{1}{\abs{S_t}} \sum_{i \in S_t} \Delta_i^t$
        \STATE Update $s_{t}$, $v_t$, and $m_t$
        \STATE ~~{\bf FedDuAdagrad}:
        \STATE ~~~~$s_t = s_{t-1} + \bar{\Delta}_t^2$, $v_t = \bar{\Delta}_t$, $m_t = \frac{1}{2\abs{S_t}} \sum_{i \in S_t} \|\Delta_i^t\|^2$
        \STATE ~~{\bf FedDuAdam}:
        \STATE ~~~~$s_t = \beta_2 s_{t-1} + (1-\beta_2)\bar{\Delta}_t^2$, $v_t = \beta_1 v_{t-1} + (1-\beta_1)\bar{\Delta}_t $, $m_t = \frac{\beta_1}{2} m_{t-1} + \frac{1-\beta_1}{2\abs{S_t}} \sum_{i \in S_t} \|\Delta_i^t\|^2$
        \STATE Compute preconditioner $G_t = \diag(\sqrt{s_t} + \epsilon)$
        \STATE Compute global learning rate $\eta_g^{t} = \frac{m_t}{\|v_t\|^2_{G_t^{-1}} + \epsilon_g}$
        \STATE Update global model: $w_{t+1} = w_t + \eta_g^t G_t^{-1} v_{t}$
        \ENDFOR
    \end{algorithmic}
\end{algorithm}

\section{Theoretical Analysis}
In this section, we conduct detailed theoretical analysis of $\fedada$
and show that dual adaptivity is essential to achieve better performance.
\subsection{Minimax Optimality}
Here, we show that our proposed global update procedure is minimax optimal under the approximate projection condition.
\begin{theorem}[Minimax Optimality]\label{theorem:minimax}
    For a given global model $w_t$ and local updates $\Delta_i^t \in \R^d$, define $H := \{w^* \mid \text{A.P.C. is satisfied}\}$
    and worst-case distance difference $V(w) := \sup_{w^* \in H} \ab[D_{\psi_t}(w^* \mid w) - D_{\psi_t}(w^* \mid w_t)]$.
    Then, for any $\psi_t$ and $\{\Delta_i^t\}_{i=1}^M$ such that $H$ is not empty,
    there exists a unique minimizer $w_{t+1}^*$ of $V(w)$
    and it matches $w_{t+1}$ of $\fedada$ defined as in Eq.~\eqref{eq:fedada-update}.
\end{theorem}
See Appendix~\ref{appendix:proof-minimax} for the proof.
Theorem~\ref{theorem:minimax} shows our proposed global update rule performs best in the worst-case scenario.
On the other hand, existing methods with partial adaptivity such as FedOpt and FedExP are suboptimal if their update differs from that of $\fedada$, since the minimizer of $V(w)$ is unique.
Thus, adaptivity to both client and gradient heterogeneity is provably necessary to update the global model optimally.

\subsection{Convergence Analysis}
Here, we prove the convergence guarantee of $\fedada$ with general distance-generating functions under the following standard assumptions.
\begin{assumption}[$L$-smoothness and Bounded Data Heterogeneity at the optimum]\label{assumption:smoothness}
    Local loss function $F_i(w)$ is differentiable and $L$-smooth.
    That is, for all $w, w' \in \R^d$, $\norm{\nabla F_i(w) - \nabla F_i(w')}_2 \leq L \norm{w - w'}_2$.
    In addition, the norm of the local gradient at the optimum $w^*$ is bounded as $\frac{1}{M} \sum_{i=1}^M \norm{\grad F_i(w^*)}^2 \leq \sigma_*^2$
\end{assumption}
\begin{theorem}\label{theorem:convergence}
    Assume that Assumption~\ref{assumption:smoothness} holds, $\{F_i\}_{i=1}^M$ are convex, and clients use full-batch SGD and participate in every round.
    Then, if $\eta_l \leq \frac{1}{6\tau L}$, $\{w^t\}_{t=1}^T$ generated by $\fedada$ satisfies
    \pagebreak
    \begin{align*}
        F\ab(\bar w_T) - F(w^*) & = \underbrace{O\ab(\frac{D_{\psi_0}(w^* \mid w_0) + \sum_{t=1}^{T-1} (D_{\psi_t}(w^* \mid w_t) - D_{\psi_{t-1}}(w^* \mid w_t))}{\sum_{t=0}^{T-1} \eta_g^t\eta_l \tau})}_{:=T_1} \\
                                & \quad + \underbrace{O(\eta_l \tau \sigma_*^2)}_{:=T_2} + \underbrace{O(\eta_l^2 \tau (\tau - 1)L\sigma_*^2)}_{:=T_3},
    \end{align*}
    where $\bar w_T = \frac{\sum_{t=0}^{T-1}\eta_g^t w_t}{\sum_{t=0}^{T-1} \eta_g^t}$.
\end{theorem}
See Appendix~\ref{appendix:proof-convergence} for the proof.
Letting $\psi_t(w) = \frac{1}{2}\norm{w}^2$ recovers the result with $T_1 = \frac{\norm{w_0 - w^*}^2}{\sum_{t=0}^{T-1} \eta_g^t\eta_l \tau}$ in~\citet{jhunjhunwala2023fedexp}
since the telescoping sum $\sum_{t=1}^{T-1} (D_{\psi_t}(w^* \mid w_t) - D_{\psi_{t-1}}(w^* \mid w_t))$ vanishes as $\psi_t=\psi_{t-1}$.
The difficulty of the proof lies in the fact that the global learning rate $\eta_g^t$ does not have a closed form solution in general, which is in contrast to the case of FedExP.
To overcome this issue, we leverage the duality and the convexity of Bregman divergence and bound the improvement at each round.
Interestingly, the bias terms $T_2$ and $T_3$ introduced by the heterogeneity are independent of the choice of $\psi_t$.
This is because $\fedada$ adaptively selects the global learning rate based on the geometry induced by $\psi_t$.
The choice of $\psi_t$ only affects the initialization error term $T_1$. Appropriate choice of $\psi_t$ reduces the numerator and improves the convergence rate.

Our theoretical analysis excludes the stochasticity by assuming full-batch SGD and full participation of clients
since theoretical analysis becomes more complicated if the global learning rate is stochastic,
as discussed in~\citet{jhunjhunwala2023fedexp}.
However, these assumptions are made solely for theoretical analysis
and we empirically demonstrate in numerical experiments that
$\fedada$ performs effectively with stochastic local updates and partial client participation.
\subsubsection{Benefit of Adaptivity} \label{section:adaptivity}
As a corollary of Theorem~\ref{theorem:convergence}, we can derive the convergence rate of $\fedadaadagrad$.
\begin{corollary}\label{corollary:convergence-adagrad}
    Assume the same conditions as Theorem~\ref{theorem:convergence} and $\sup_{t=0}^T \norm{w_t - w^*}_{\infty} \leq D$.
    Then, $\{w^t\}_{t=1}^T$ generated by FedDuAdagrad satisfies
    \begin{align*}
        F\ab(\bar w_T) - F(w^*) = O\ab(\frac{D^2 \tr(G_{T-1})}{\sum_{t=0}^{T-1} \eta_g^t\eta_l \tau}) + O(\eta_l \tau \sigma_*^2) + O(\eta_l^2 \tau (\tau - 1)L\sigma_*^2)
    \end{align*}
\end{corollary}
See Appendix~\ref{appendix:proof-convergence-adagrad} for the proof.

To see the benefit of the adaptivity,
let us consider the case where the local update is anisotropic, which is often the case in modern machine learning tasks~\citep{faghri2020study,tomihari2025understanding}.
Specifically, we assume $\abs{[\bar \Delta_t]_{k}} = \Theta(a_t \cdot k^{-\beta})$ for some $a_t > 0, \beta > 1$.
That is, the magnitude of the $k$-th element of the local update decays polynomially.
Then, if $\epsilon, \epsilon_g = 0$, the initialization error term $T_1$ can be evaluated as
\begin{align*}
    T_1 = \frac{D^2 \tr(G_{T-1})}{\sum_t \eta_g^t\eta_l \tau} = O\ab(\frac{D^2\sqrt{\sum_{t=0}^{T-1} a_t^2}}{\eta_l \tau \cdot \sum_{t=0}^{T-1} \sqrt{\sum_{s=0}^t a_s^2}}).
\end{align*}
As $w_t$ approaches the optimal solution, it is reasonable to assume that the magnitude of the averaged local update $a_t$ decreases.
Then, if $a_t$ is monotonically decreasing, we obtain $T_1 = O(D^2 / (T\eta_l \tau))$.
See Appendix~\ref{appendix:adaptivity} for the detailed derivation.
Thus, the convergence rate of $\fedadaadagrad$ is independent of the dimension $d$.
This is in contrast to the case of FedExP and FedAvg, where the convergence rate $O\ab(\frac{\norm{w_0 - w^*}^2}{\sum_{t=0}^{T-1} \eta_g^t \eta_l \tau}) = O\ab(\frac{dD^2}{\sum_{t=0}^{T-1} \eta_g^t \eta_l \tau})$
is linearly dependent on $d$ since $\norm{w_0 - w_*}^2 = O(dD^2)$.
Thus, if $d \gg 1$, $\fedadaadagrad$ is expected to converge faster than FedExP and FedAvg.

\section{Numerical Experiments}

We evaluate the performance of $\fedada$ on synthetic and real-world datasets.
We consider a distributed overparameterized linear regression problem for synthetic datasets, and image classification and NLP tasks for real-world datasets.
We compare $\fedada$ with the following baselines: FedAvg, FedExP, FedOpt (FedAdagrad and FedAdam), and their momentum variants~(FedAvgM, FedExPM).
These algorithms do not require additional computational or communication cost on clients as our proposed method.
As discussed in~\citet{jhunjhunwala2023fedexp}, adaptive learning rate can cause oscillating behavior in the performance.
Thus, we adopt averaging the last two iterates strategy as in~\citet{jhunjhunwala2023fedexp}.
For a fair comparison, we perform grid-search over $\eta_g, \eta_l$ for FedAvg and FedOpt,
and $\epsilon_g, \eta_l$ for FedExP and $\fedada$. We fix $\beta_1 = 0.9$ and $\beta_2 = 0.99$ for Fed(Du)Adam following~\citet{reddi2021adaptive}
and set $\epsilon = 10^{-9}$ for FedOpt and $\fedada$ if not specified.
We fix the number of participating clients at each round to $20$, minibatch size for local SGD to $50$, and the number of local updates to $\tau = 20$.
The results are averaged over 5 different random seeds and the shade represents the standard deviation.
See Appendix~\ref{appendix:experimental-setup} for the detailed experimental setup.

\begin{table}[t]
    \centering
    \caption{Average validation accuracy (\%) of the last iterate over 5 different random seeds.
        Results within 0.5\% of the best result for each dataset are bolded.}
    \label{table:results}
    \begin{tabular}{ccccccccc}
        \toprule
        Fed...      & DuAdam        & DuAdagrad     & ExP  & ExPM          & Adam & Adagrad & AvgM & Avg  \\
        dataset     & (ours)        & (ours)        &      &               &      &         &      &      \\
        \midrule
        CIFAR100    & \textbf{62.9} & 51.2          & 48.4 & 59.8          & 58.4 & 40.9    & 56.5 & 39.5 \\
        CIFAR10     & \textbf{86.6} & 82.1          & 80.7 & \textbf{86.4} & 81.9 & 74.0    & 80.6 & 73.1 \\
        FEMNIST     & \textbf{78.0} & \textbf{78.3} & 77.5 & 75.8          & 77.2 & 71.6    & 76.0 & 76.6 \\
        shakespeare & 50.9          & \textbf{52.6} & 50.6 & 48.9          & 50.6 & 51.0    & 48.4 & 49.1 \\
        \bottomrule
    \end{tabular}
    \vspace{-0.5em}
\end{table}

\begin{figure}[t]
    \centering
    \includegraphics[width=0.99\textwidth]{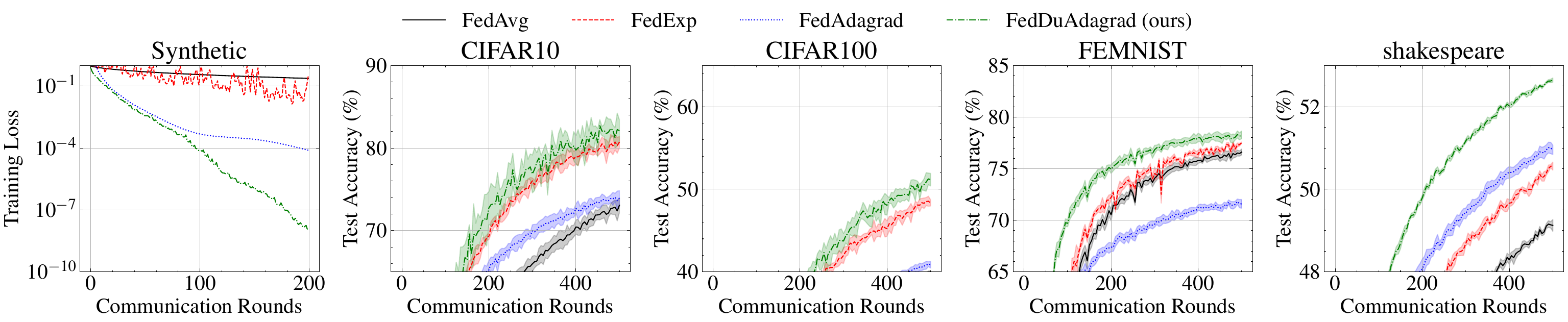}
    \includegraphics[width=0.99\textwidth]{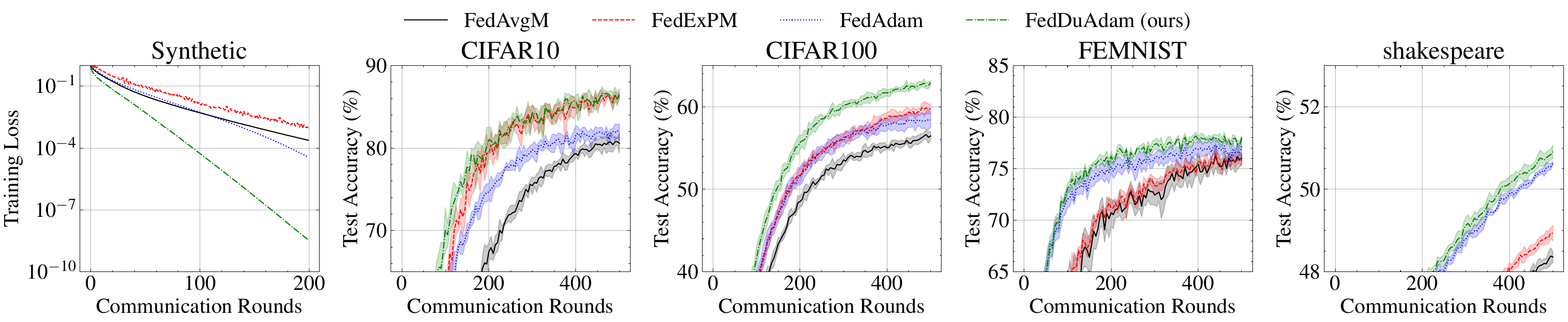}
    \caption{Test accuracy for $\fedada$ and baselines without server momentum (upper)
        and with server momentum (lower).
        Our proposed methods (green dashdot) consistently outperform baselines.}
    \label{fig:experiment_images}
    \vspace{-0.5em}
\end{figure}

\begin{figure}[t]
    \centering
    \begin{minipage}[t]{0.74\textwidth}
        \centering
        \includegraphics[width=0.32\textwidth]{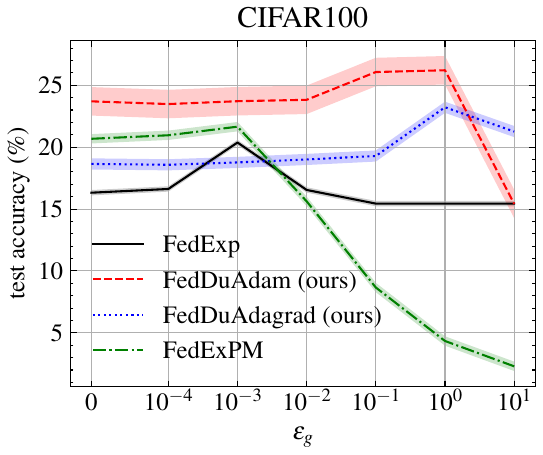}
        \includegraphics[width=0.32\textwidth]{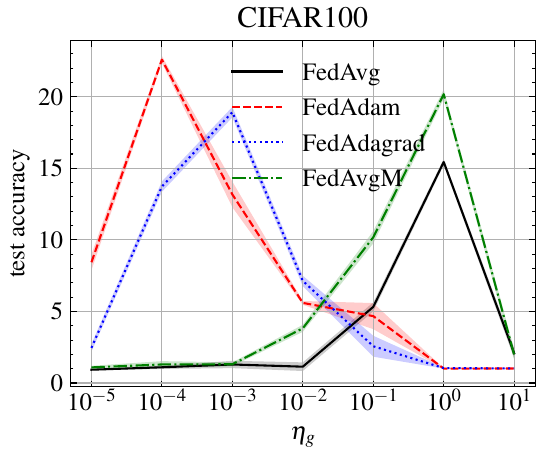}
        \includegraphics[width=0.32\textwidth]{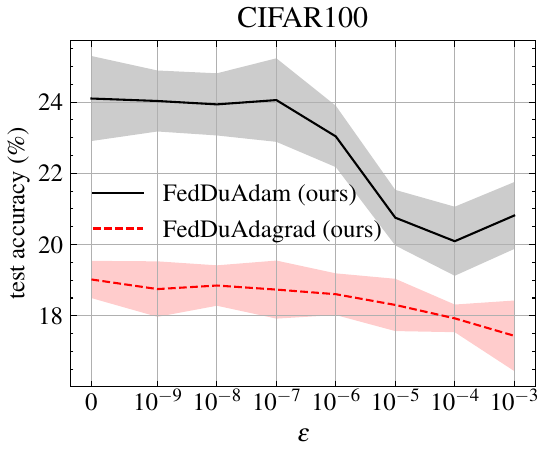}
        \caption{Test accuracy averaged over the last 5 iterates with different hyperparameters $\epsilon_g, \eta_g$ and $\epsilon$. Proposed methods are less sensitive to the choice of hyperparameters.}
        \label{fig:tuning}
    \end{minipage}
    \hfill
    \begin{minipage}[t]{0.24\textwidth}
        \centering
        \includegraphics[width=0.96\textwidth]{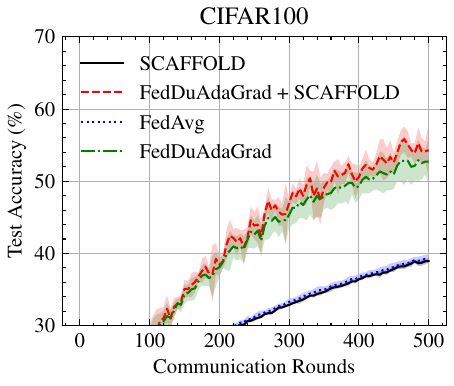}
        \caption{Performance of $\fedada$ with SCAFFOLD.}
        \label{fig:scaffold}
    \end{minipage}
    \vspace{-0.5em}
\end{figure}

\paragraph{Synthetic Datasets}
For the synthetic experiment, we generate $M = 20$ clients with $\abs{\mathcal{D}_i} = 30$ samples.
The synthetic datasets are generated following a similar procedure as in~\citet{jhunjhunwala2023fedexp} and~\citet{li2020federated},
but we consider anisotropic data distribution, which corresponds to the fact that data often lies on a low-dimensional structure~\citep{ansuini2019intrinsic}.
Specifically, we generate the input data $x \in \R^d~(d=1000)$ as $x \sim \mathcal{N}(0, \Sigma)$, where $\Sigma$ is a diagonal matrix with its $k$-th diagonal element $\Sigma_{kk} = k^{-\beta}$ for $\beta = 1.1$.

\paragraph{Real-world Datasets}
For CIFAR10/100, we parition the data into $M=100$ clients by following a Dirichlet distribution with parameter $\alpha = 0.3$
and use ResNet-18 architecture.
For FEMNIST, data is naturally partitioned into $3,550$ clients based on the writer of the digit or character~\citep{caldas2018leaf}.
We subsample $100$ clients for train and test to reduce the computational cost, and use the same CNN architecture as in~\citet{zhu2022diurnal}.
Shakespeare dataset is also naturally partitioned into $1,129$ clients based on the speaking role~\citep{caldas2018leaf}
and we subsample $100$ clients and use 1-layer LSTM for next character prediction.

\paragraph{$\fedada$ consistently outperforms baselines}
Fig.~\ref{fig:experiment_images} and Table~\ref{table:results} show the performance of $\fedada$ and baselines on synthetic and real-world datasets.
Overall, $\fedada$ consistently outperforms existing methods across all datasets.
These results clearly show that dual adaptivity is a key to achieve fast convergence in FL.
In CIFAR10/100, momentum-based methods perform better than non-momentum methods
while non-momentum methods perform better or comparable in the other datasets.
See Appendix~\ref{appendix:long-term} for long-term behavior of the algorithms.

\paragraph{$\fedada$ is Robust to the choice of hyperparameters}
Hyperparameter-tuning is often time-consuming and expensive in FL.
Adaptive methods are expected to be robust to the choice of hyperparameters.
To see this, we run each algorithm with different choice of $\epsilon_g, \eta_g$ for 50 rounds.
We tune $\eta_l$ for each hyperparameter setting.
As shown in Fig.~\ref{fig:tuning}, the performance of $\fedada$ is not sensitive to the choice of $\epsilon_g$.
Furthermore, $\epsilon_g = 0$ is sufficient to achieve good performance.
This indicates that $\fedada$ is robust to the choice of hyperparameters and
even hyperparameter-free by setting $\eta_g^t := m_t/ \norm{v_t}^2_{G_t^{-1}}$.
On the other hand, FedOpt does not perform well with not well-tuned hyperparameters,
which requires careful tuning of $\eta_g$.
We also conduct an ablation study on the choice of $\epsilon$ for $\fedada$.
Note that $\epsilon$ determines the degree of adaptivity on gradient heterogeneity.
We see that $\fedada$ is also robust to the choice of $\epsilon$ but increasing $\epsilon$ degrades the performance gradually since it reduces the gradient adaptivity.
We find that $\epsilon = 0$ works well but we use $\epsilon = 10^{-9}$ in other experiments for the sake of numerical stability.

\paragraph{SCAFFOLD with dual adaptivity}
Since our approach is orthogonal to existing methods which modify the local update procedure such as SCAFFOLD~\citep{karimireddy2020scaffold},
we can combine our method with them to further improve the performance.
We compare in Fig.~\ref{fig:scaffold} the performance of vanilla SCAFFOLD and $\fedada$ with SCAFFOLD-type local update procedure.
We see that vanilla SCAFFOLD does not outperform FedAvg, which is consistent with the results in~\citet{jhunjhunwala2023fedexp}, but $\fedada$ with SCAFFOLD outperforms the other methods.

\begin{wrapfigure}{r}[0pt]{0.25\textwidth}
    \centering
    \vspace{-1.5em}
    \includegraphics[width=0.24\textwidth]{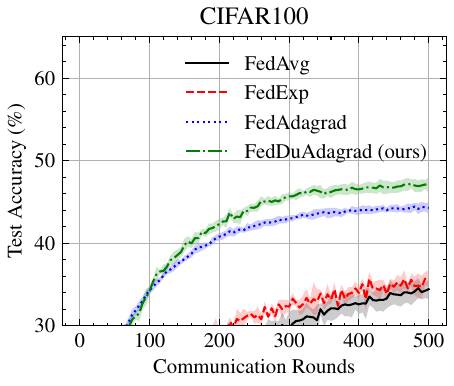}
    \caption{Training of ViT on CIFAR-100.}
    \label{fig:transformer}
    \vspace{-1em}
\end{wrapfigure}

\paragraph{Coordinate-wise adaptivity is essential in training of Transformers}
As discussed in previous works~\citep{zhang2024transformers,tomihari2025understanding},
adaptive methods such as Adam outperform SGD in centralized training of Transformers.
To see the effect of adaptivity in FL, we train a Vision Transformer (ViT)~\citep{dosovitskiy2021image} on CIFAR-100.
Fig.~\ref{fig:transformer} shows that FedExP does not work well in training of ViT while it performs comparably in training of ResNet.
This is in contrast to $\fedadaadagrad$, which performs well in both cases. The result implies that coordinate-wise adaptivity is also effective in FL training of Transformers.

\section{Conclusion}
In this paper, we addressed the question of how to design a global update procedure for FL, which is adaptive to both client and gradient heterogeneity.
For this purpose, we formulated the global update procedure through the lens of mirror descent
and proposed $\fedada$, a doubly adaptive step-size rule for general mirror descent-type algorithms.
We proved that our proposed step-size is minimax optimal under approximate projection condition and provided the convergence analysis for convex objectives.
Extensive numerical experiments show that $\fedada$ outperforms existing adaptive methods in various settings without requiring additional computational and communication cost on clients.
Furthermore, $\fedada$ is robust to the choice of hyperparameters and can be combined with existing methods such as SCAFFOLD to further improve the performance.
An interesting future direction is to extend our theoretical analysis to non-convex objectives and partial client participation.
\bibliography{ref}

\bibliographystyle{abbrvnat}
\appendix


\section{Proof for Theorem~\ref{theorem:lower-bound}}\label{appendix:proof-lower-bound}
The approximate projection condition yields
\begin{align*}
    \langle \bar \Delta_t, w^* - w_t\rangle \geq \frac{1}{2M} \sum_{i=1}^M \|\Delta_i^{t}\|^2 \geq 0.
\end{align*}
For $\theta_{t+1} := \theta_t + \eta_g \bar \Delta_t$, we have
\begin{align*}
    D_{\phi_t}(\theta_{t+1} \mid \theta^*) - D_{\phi_t}(\theta_t \mid \theta^*) & = \phi_t(\theta_{t+1}) - \phi_t(\theta_t) - \eta_g \langle \nabla \phi_t(\theta^*), \bar \Delta_t \rangle \\
                                                                                & = \phi_t(\theta_t + \eta_g \bar \Delta_t) - \phi(\theta_t) - \eta_g \langle w^*, \bar \Delta_t \rangle.
\end{align*}
For the last inequality, we used $\grad \phi(\theta^*) = w^*$ from the duality.
From the first-order optimality condition, we have
\begin{align*}
    h_t(\eta_g) = \langle w^* - w_t, \bar \Delta_t \rangle,
\end{align*}
where $h_t(\eta) = \odv{}{\eta} \phi(\theta_t + \eta \bar \Delta_t) - \lrangle{w_t, \bar \Delta_t}$.
Since $\phi_t$ is assumed to be strongly convex, there exists a constant $\alpha_t > 0$ such that $\grad^2 \phi(\theta) \succ \alpha_t I_d$,
and we have
\begin{align*}
    \odv[1]{h_t}{\eta}(\eta) & = \langle \nabla^2 \phi(\theta_t + \eta \bar \Delta_t) \bar \Delta_t, \bar \Delta_t\rangle > \alpha_t \norm{\bar \Delta_t}^2.
\end{align*}
This implies that $h_t$ is strictly increasing and the inverse function $h_t^{-1}$ exists on $\R$.
Thus, $\eta_g^*$ is uniquely determined by
\begin{align*}
    \eta_g^* = h_t^{-1}\left(\langle w^* - w_t, \bar \Delta_t \rangle \right).
\end{align*}

Using the monotonicity of $h_t^{-1}$, we have
\begin{align*}
    \eta_g^* & = h_t^{-1}(\langle w^* - w_t, \bar \Delta_t \rangle)                  \\
             & \geq h_t^{-1}\left(\frac{1}{2M}\sum_{i=1}^M \|\Delta_i^t\|^2 \right),
\end{align*}
which completes the proof.

\section{Proof for Theorem~\ref{theorem:lower-bound-momentum}}\label{appendix:proof-lower-bound-momentum}
The strong approximate projection condition yields
\begin{align*}
    \langle \bar \Delta_t, w^* - w_t\rangle \geq \frac{1}{M} \sum_{i=1}^M \|\Delta_i^{t}\|^2 \geq 0.
\end{align*}
For $\theta_{t+1} := \theta_t + \eta_g v_t$, we have
\begin{align*}
    D_{\phi_t}(\theta_{t+1} \mid \theta^*) - D_{\phi_t}(\theta_{t} \mid \theta^*) & = \phi_t(\theta_{t+1}) - \phi_t(\theta_{t}) - \eta_g \langle \nabla \phi_t(\theta^*), v_{t} \rangle \\
                                                                                  & = \phi_t(\theta_{t} + \eta_g v_{t}) - \phi_t(\theta_{t}) - \eta_g \langle w^*, v_t \rangle.
\end{align*}
For the last inequality, we used $\grad \phi(\theta^*) = w^*$ from the duality.
From the first-order optimality condition, we have
\begin{align*}
    h_t(\eta_g) & = \langle w^* - w_{t}, v_{t} \rangle.
\end{align*}
and $\eta_g^* = h_t^{-1}(\langle w^* - w_t, v_t \rangle)$
as in the proof of Theorem~\ref{theorem:lower-bound}.
From the monotonicity of $h_t^{-1}$, it suffices to show
\begin{align*}
    \langle w^* - w_t, v_t \rangle \geq 2m_t \geq m_t.
\end{align*}
To prove this, we use induction on $t$.
\paragraph{Case $t=0$}
We have
\begin{align*}
    \langle w^* - w_0, v_0 \rangle & = (1-\beta_1)\langle w^* - w_0, \bar \Delta_0\rangle                 \\
                                   & \geq (1-\beta_1) \frac{1}{M}\sum_{i=1}^M \norm{\Delta_i^0}^2 = 2m_0.
\end{align*}
The inequality follows from strong A.P.C.
Thus, we obtain the desired result.
\paragraph{Case $t \geq 1$}
Assume that the inequality holds for $t-1$.
Then, we have
\begin{align*}
    \langle w^* - w_t, v_t \rangle & = (1-\beta_1)\langle w^* - w_t, \bar \Delta_t\rangle + \beta_1 \langle w^*-w_t, v_{t-1}\rangle                                                                     \\
                                   & = (1-\beta_1)\langle w^* - w_{t}, \bar \Delta_t\rangle + \beta_1 (\langle w^*-w_{t-1}, v_{t-1}\rangle - \underbrace{\langle w_t - w_{t-1}, v_{t-1}\rangle}_{:=R}).
\end{align*}
The term $R$ can be evaluated as
\begin{align*}
    R & = \langle \grad \phi_{t-1}(\theta_{t-1} + \eta_g^{t-1}v_{t-1}), v_{t-1} \rangle - \langle w_{t-1}, v_{t-1}\rangle \\
      & = h_{t-1}(\eta_g^{t-1}) \leq m_{t-1}
\end{align*}
since $\eta_g^{t-1}$ is assumed to be smaller than $h^{-1}_{t-1}(m_{t-1})$.
Substituting the above equality, we obtain
\begin{align*}
    \langle w^* - w_t, v_t \rangle & = (1-\beta_1)\langle w^* - w_t, \bar \Delta_t\rangle + \beta_1 (\langle w^*-w_{t-1}, v_{t-1}\rangle - m_{t-1}) \\
                                   & \geq (1-\beta_1)\langle w^* - w_t, \bar \Delta_t\rangle + \beta_1 (2m_{t-1} - m_{t-1})                         \\
                                   & \geq (1-\beta_1)\frac{1}{M}\sum_{i=1}^M \norm{\Delta_i^t}^2 + \beta_1 m_{t-1}                                  \\
                                   & = 2m_{t}.
\end{align*}
Here, we used the induction hypothesis for the first inequality and strong A.P.C. for the second inequality.
Then, we obtain the result by induction.

\section{Proof for Theorem~\ref{theorem:minimax}}\label{appendix:proof-minimax}
Given $w \in \R^d$, the worst-case distance difference is defined as
\begin{align*}
    V(w) & = \sup_{w^* \in H} V(w, w^*)                                          \\
         & := \sup_{w^* \in H} D_{\psi_t}(w^*\mid w) - D_{\psi_t}(w^* \mid w_t).
\end{align*}
From the definition of the Bregman divergence, we have
\begin{align*}
    V(w, w^*) & = D_{\psi_t}(w^*\mid w) - D_{\psi_t}(w^* \mid w_t)                                                \\
              & = - \psi_t(w) - \grad \psi_t(w)^\top (w^* - w) + \psi_t(w_t) + \grad \psi_t(w_t)^\top (w^* - w_t) \\
              & = \psi_t(w_t) - \psi_t(w) + (\grad \psi_t(w_t)- \grad \psi_t(w))^\top (w^* - w).
\end{align*}
Thus, $V(w, w^*)$ is affine in $w^*$.

Let us consider the Lagrangian
\begin{align*}
    \mathcal{L}(w, w^*, \lambda) = V(w, w^*) - \lambda \ab(\langle \bar \Delta^t, w^t - w^*\rangle + \frac{1}{2M}\sum_{i=1}^M \norm{\Delta_i^t}^2)
\end{align*}
Since $V(w, w^*)$ and the constraint are affine,
strong duality holds and we have
\begin{align*}
    V(w) & = \sup_{w^* \in \R^d}\inf_{\lambda \geq 0} \mathcal{L}(w, w^*, \lambda)   \\
         & = \inf_{\lambda \geq 0} \sup_{w^* \in \R^d} \mathcal{L}(w, w^*, \lambda),
\end{align*}
and
\begin{align*}
    \inf_{w} V(w)
     & = \inf_{w} \inf_{\lambda \geq 0} \sup_{w^* \in \R^d} \mathcal{L}(w, w^*, \lambda)  \\
     & = \inf_{\lambda \geq 0} \inf_{w} \sup_{w^* \in \R^d} \mathcal{L}(w, w^*, \lambda).
\end{align*}
Since the Lagrangian is affine, $\sup_{w^* \in \R^d} \mathcal{L}(w, w^*, \lambda)$ is infinite unless $\theta_t - \theta + \lambda \bar \Delta_t = 0$
and if this condition holds, the value of $\sup_{w^* \in \R^d} \mathcal{L}(w, w^*, \lambda)$ is independent of $w^*$.
Thus, we have
\begin{align*}
    \sup_{w^* \in \R^d} \mathcal{L}(w, w^*, \lambda) = \begin{cases}
                                                           V(w, w_t) + \frac{\lambda}{2M}\sum_{i=1}^M \norm{\Delta_i^t}^2 & \text{if } \theta_t - \theta + \lambda \bar \Delta_t = 0, \\
                                                           \infty                                                         & \text{otherwise}.
                                                       \end{cases}
\end{align*}
Therefore, for a given $\lambda\geq 0$, $\sup_{w^*}\mathcal{L}(w, w^*, \lambda)$ is minimized at $w = w^\lambda := \grad \phi_t(\theta_t + \lambda \bar \Delta_t)$
Therefore, we have
\begin{align*}
    \inf_{w} \sup_{w^* \in \R^d} \mathcal{L}(w, w^*, \lambda) & = V(w^\lambda, w_t) + \frac{\lambda}{2M}\sum_{i=1}^M \norm{\Delta_i^t}^2                                                                                                    \\
                                                              & = D_{\psi_t}(w_t \mid w^\lambda) - \frac{\lambda}{2M}\sum_{i=1}^M \norm{\Delta_i^t}^2                                                                                       \\
                                                              & = D_{\phi_t}(\theta_t + \lambda \bar \Delta_t \mid \theta_t) - \frac{\lambda}{2M}\sum_{i=1}^M \norm{\Delta_i^t}^2                                                           \\
                                                              & = \phi(\theta_t + \lambda \bar \Delta_t) - \phi(\theta_t) - \langle \grad \phi(\theta_t),\lambda \bar \Delta_t\rangle - \frac{\lambda}{2M}\sum_{i=1}^M \norm{\Delta_i^t}^2.
\end{align*}
Considering the first-order optimality condition on $\lambda$, we have
\begin{align}
     & \odv{}{\lambda} \phi(\theta_t + \lambda \bar \Delta_t) - \langle \grad \phi(\theta_t), \bar \Delta_t \rangle - \frac{1}{2M}\sum_{i=1}^M \norm{\Delta_i^t}^2 \label{eq:diff-phi} \\
     & \quad = \odv{}{\lambda} \phi(\theta_t + \lambda \bar \Delta_t) - \langle w_t, \bar \Delta_t \rangle - \frac{1}{2M}\sum_{i=1}^M \norm{\Delta_i^t}^2          \notag              \\
     & \quad = h_t(\lambda)- \frac{1}{2M}\sum_{i=1}^M \norm{\Delta_i^t}^2,\notag                                                                                                       \\
     & \quad = 0,\notag
\end{align}
which implies the optimal $\lambda_* = h_t^{-1}(\frac{1}{2M}\sum_{i=1}^M \norm{\Delta_i^t}^2) = \eta_g^t$ uniquely exists as in the proof of Theorem~\ref{theorem:lower-bound}.
Note that $\eta_g^t \geq 0$ since expression \eqref{eq:diff-phi} is increasing in $\lambda$ and negative at $\lambda = 0$.
Thus, $V(w)$ is minimized at $w = \grad \phi(\theta_t + \eta_g^t \bar \Delta_t)$.
This completes the proof.

\section{Proof for Theorem~\ref{theorem:convergence}}\label{appendix:proof-convergence}
Let $D_t = D_{\psi_t}$ for simplicity.
We have
\begin{align*}
    D_t(w^* \mid w_{t+1}) - D_t(w^* \mid w_t) & = D_{\phi_t}(\theta_{t+1} \mid \theta^*) - D_{\phi_t}(\theta_t \mid \theta^*)                           \\
                                              & = \phi_t(\theta_{t+1}) - \phi_t(\theta_t) - \lrangle{\grad \phi_t(\theta^*), \theta_{t + 1} - \theta_t} \\
                                              & = \phi_t(\theta_{t+1}) - \phi_t(\theta_t) - \eta_g^t \lrangle{w^*, \bar \Delta_t}                       \\
                                              & = H_t(\eta_g^t) + \eta_g^t \lrangle{w_t - w^*, \bar \Delta_t},
\end{align*}
where $H_t(\eta) = \phi_t(\theta_t + \eta \bar \Delta_t) - \phi_t(\theta_t) - \eta \lrangle{w_t, \bar \Delta_t}$.
For the second equality, we used the duality $w^* = \grad \phi_t(\theta^*)$.
From the mean-value theorem, there exists $\bar \eta \in [0, \eta_g^t]$ such that
\begin{align*}
    \frac{H_t(\eta_g^t) - H_t(0)}{\eta_g^t - 0} = \frac{H_t(\eta_g^t)}{\eta_g^t} = \odv{H_t}{\eta}(\bar \eta) = h_t(\bar \eta) \leq h_t(\eta_g^t).
\end{align*}
The last inequality follows from the fact that $h_t$ is increasing and $\bar \eta \leq \eta_g^t$.
From the definition of $\eta_g^t$, we have
\begin{align*}
    \frac{H_t(\eta_g^t)}{\eta_g^t} \leq h_t(\bar \eta_g^t) = \frac{1}{2M} \sum_{i=1}^M \norm{\Delta_i^t}^2.
\end{align*}
Substituting the above inequality, we have
\begin{align*}
    D_t(w^* \mid w_{t + 1}) - D_t(w^* \mid w_{t + 1})
     & \leq \frac{\eta_g^t}{2} \cdot \underbrace{\frac{1}{M} \sum_{i=1}^M \norm{\Delta_i^t}^2}_{:=R_2} + \eta_g^t \underbrace{\lrangle{w_t - w^*, \bar \Delta_t}}_{:=R_1}.
\end{align*}
In a similar way as in the proof of Theorem 1 in~\citet{jhunjhunwala2023fedexp},
$R_2$ can be bounded as
\begin{align*}
    R_2 & = \frac{1}{M} \sum_{i=1}^M \norm{\Delta^t_i}^2                                                                                                                \\
        & = \frac{\eta_l^2}{M} \sum_{i=1}^M \norm{\sum_{k=0}^{\tau - 1} \grad F_i(w^t_{i,k})}^2                                                                         \\
        & \leq \frac{\tau \eta_l^2}{M} \sum_{i=1}^M \sum_{k=0}^{\tau - 1} \norm{\grad F_i(w^t_{i,k})}^2                                                                 \\
        & \leq \frac{3\tau\eta_l^2 L^2}{M}\sum_{i=1}^M \sum_{k=0}^{\tau - 1} \norm{w^t_{i,k} - w_t}^2 + 6\tau^2\eta_l^2 L(F(w^t) - F(w^*)) + 3\tau^2\eta_l^2\sigma_*^2.
\end{align*}
Here, we used Lemma 5 in~\citet{jhunjhunwala2023fedexp} for the last inequality.
For $R_1$, we have
\begin{align*}
    R_1 & = \frac{1}{M} \sum_{i=1}^M \langle w_t - w^*, \Delta^t_i \rangle                                        \\
        & = \frac{\eta_l}{M} \sum_{i=1}^M \sum_{k=0}^{\tau - 1} \langle w_t - w^*, \grad F_i(w_{i, k}^t) \rangle.
\end{align*}
As shown in the proof of Theorem 1 in~\citet{jhunjhunwala2023fedexp},
the right-hand side can be bounded as
\begin{align*}
    R_1 & \geq \eta_l \tau (F(w_t) - F(w^*)) - \frac{\eta_l L}{2M} \sum_{i=1}^M \sum_{k=0}^{\tau - 1} \norm{w^t_{i,k} - w_t}^2.
\end{align*}

Combining the above two inequalities, we have
\begin{align*}
    D_t(w^* \mid w_{t + 1}) - D_t(w^* \mid w_t) & \leq 2\eta_g^t \eta_l \tau(1 - 3\eta_l \tau L) (F(w_t) - F(w^*)) + 3 \eta_g^t \eta_l^2\tau^2 \sigma_*^2                                     \\
                                                & +(3 \eta_g^t \eta_l^2\tau + \eta_g^t \eta_l L) \frac{1}{M} \sum_{i=1}^M \sum_{k=0}^{\tau - 1} \norm{w^t_{i,k} - w_t}^2                      \\
                                                & \leq -\frac{\eta_g^{t}\eta_l \tau}{3} (F(w_t) - F^*) + \eta_g^{t} \cdot O(\eta_l^3\tau^2\sigma_*^2 + \eta_l^2 \tau^2(\tau - 1)L\sigma_*^2).
\end{align*}

Averaging over all rounds, we have
\begin{align*}
    \frac{\sum_{t=0}^{T-1} \eta_g^t (F(w_t) - F(w^*))}{\sum_{t=0}^{T-1} \eta_g^t} & \leq 3\cdot \frac{\sum_{t=0}^{T-1} D_t(w^* \mid w_t) - D_t(w^* \mid w_{t+1})}{\sum_t \eta_g^t\eta_l \tau} + O(\eta_l \tau \sigma_*^2 + \eta_l^2 \tau (\tau - 1)L\sigma_*^2) \\
                                                                                  & \leq O\ab(\frac{D_0(w^* \mid w_0) + \sum_{t=1}^{T-1} (D_t(w^* \mid w_t) - D_{t-1}(w^* \mid w_t))}{\sum_t \eta_g^t\eta_l \tau})                                              \\
                                                                                  & \quad + O(\eta_l \tau \sigma_*^2 + \eta_l^2 \tau (\tau - 1)L\sigma_*^2).
\end{align*}
Since $F$ is convex, we have
\begin{align*}
    F(\bar w_T) - F(w^*) & \leq \frac{\sum_{t=0}^{T-1} \eta_g^t (F(w_t) - F(w^*))}{\sum_{t=0}^{T-1} \eta_g^t}.
\end{align*}
This completes the proof.

\section{Proof for Corollary~\ref{corollary:convergence-adagrad}}\label{appendix:proof-convergence-adagrad}
The numerator in the first term can be bounded as
\begin{align*}
    \sum_{t = 0}^{T-1} D_t(w^* \mid w_t) - D_t(w^* \mid w_{t + 1}) & \leq \sum_{t=0}^{T-1} (D_t(w^* \mid w_t) - D_{t-1}(w^* \mid w_t))                     \\
                                                                   & = \sum_{t = 0}^{T-1}\frac{1}{2} (w_t - w^*)^\top (G_{t} - G_{t-1})(w_t - w^*)         \\
                                                                   & \leq \sum_{t = 0}^{T-1}\frac{1}{2} \norm{w_t - w^*}_\infty^2 \norm{g_{t} - g_{t-1}}_1 \\
                                                                   & = \sum_{t = 0}^{T-1}\frac{D^2}{2} (\norm{g_{t}}_1 - \norm{g_{t-1}}_1)                 \\
                                                                   & \leq \frac{D^2}{2} \norm{g_{T-1}}_1                                                   \\
                                                                   & = \frac{D^2}{2} \tr(G_{T-1})
\end{align*}
where we define $D_{-1}(w_0 \mid w^*) = 0$ and $g_t = \diag(G_t)$.
The second equality follows from the monotonicity of $[g_t]_k$.
Substituting the above inequality, we obtain
\begin{align*}
    F\ab(\frac{\sum_{t=0}^{T-1} \eta_g^t w_t}{\sum_{t=0}^{T-1} \eta_g^t}) - F(w^*) & \leq \frac{\sum_{t=0}^{T-1} \eta_g^t (F(w_t) - F(w^*))}{\sum_{t=0}^{T-1} \eta_g^t}                                                         \\
                                                                                   & = O\ab(\frac{D^2 \tr(G_{T-1})}{\sum_{t=0}^{T-1} \eta_g^t\eta_l \tau}) + O(\eta_l \tau \sigma_*^2) + O(\eta_l^2 \tau (\tau - 1)L\sigma_*^2)
\end{align*}
This completes the proof.

\section{Detailed Analysis on Benefit of Adaptivity}\label{appendix:adaptivity}
From the assumption, we have
\begin{align*}
    [G_{t}]_{k,k}
     & = \sqrt{\sum_{t=0}^{t} [\bar \Delta_t]_k^2}         \\
     & = \Theta\ab(k^{-\beta}\sqrt{\sum_{s=0}^{t} a_s^2}).
\end{align*}
Thus, $\tr(G_{T-1}) = \Theta(\sqrt{\sum_{t=0}^{T-1} a_t^2})$ since $\beta > 1$.
On the other hand, the numerator can be bounded as
\begin{align*}
    \sum_t \eta_g^{(t)}
     & = \sum_{t=0}^{T-1} \frac{\frac{1}{M}\sum_{i=1}^M \norm{\Delta_i^t}^2}{\norm{\bar \Delta^t}_{G_t^{-1}}^2}                                              \\
     & \geq \sum_{t=0}^{T-1} \frac{\norm{\bar \Delta^t}^2}{\norm{\bar \Delta^t}_{G_t^{-1}}^2}                                                                \\
     & = \sum_{t=0}^{T-1} \frac{\Theta\ab(\sum_{k=1}^d a_t^2 k^{-2\beta})}{\Theta\ab(\frac{a_t^2}{\sqrt{\sum_{s=0}^t a_s^2}} \cdot \sum_{k=1}^d k^{-\beta})} \\
     & = \Theta\ab(\sum_{t=0}^{T-1} \sqrt{\sum_{s=0}^t a_s^2}),
\end{align*}
since $[G_t]_{k,k} = \sqrt{\sum_{s=0}^t [\bar \Delta_s]_k^2} = \Theta(\sqrt{\sum_{s=0}^t a_s^2 k^{-2\beta}})$.
Substituting the above two inequalities, we have
\begin{align*}
    T_1 = O\ab(\frac{D^2\sqrt{\sum_{t=0}^{T-1} a_t^2}}{\eta_l \tau \sum_{t=0}^{T-1} \sqrt{\sum_{s=0}^t a_s^2}}).
\end{align*}
If $a_t$ is decreasing, we have\begin{align*}
    \frac{D^2}{\eta_l \tau \sum_{t=0}^{T-1} \frac{\sqrt{\sum_{s=0}^t a_s^2}}{\sqrt{\sum_{s=0}^{T-1} a_s^2}}}
     & \leq \frac{D^2}{\eta_l \tau \sum_{t=0}^{T-1} \sqrt{\frac{t}{T}}} \\
     & = O\ab(\frac{D^2}{\eta_l \tau T}).
\end{align*}
For the first inequality, we used $\frac{\sum_{s=0}^t a_s^2}{\sum_{s=0}^{T-1} a_s^2} \geq t / T$ since we assume $a_{s} \leq a_{s-1}$.
That is, we have
\begin{align*}
    \frac{\sum_{s=0}^{T-1} a_s^2}{\sum_{s=0}^t a_s^2} & = 1 + \frac{\sum_{s=t+1}^{T-1} a_s^2}{\sum_{s=0}^{t} a_s^2} \\
                                                      & \leq 1 + \frac{(T-t-1) \cdot a_{t}^2}{(t+1)\cdot a_t^2}     \\
                                                      & = \frac{T}{t+1}
\end{align*}
and thus $\frac{\sum_{s=0}^t a_s^2}{\sum_{s=0}^{T-1} a_s^2} \geq \frac{t}{T}$.
\section{Detailed Experimental Setup}\label{appendix:experimental-setup}
\subsection{Compute Resources and Time}~\label{appendix:compute-resources}
Our experiments were conducted on Intel(R) Xeon(R) Silver 4316 CPU @ 2.30GHz and 8 NVIDIA A100-SXM4-80GB GPUs.
The training process (500 rounds) takes about 3 hours for each method and dataset.

\subsection{Dataset and Model}
We summarize the datasets and models used in our main experiments in Table~\ref{tab:datasets-models}.
We also provide the architecture of LSTM and CNN in Table~\ref{tab:lstm_architecture} and Table~\ref{tab:cnn_architecture}, respectively.
For ViT experiments, we use the architecture in~\citet{omiita2024vit}.
\begin{table}[h]
    \centering
    \caption{Datasets and models used in our experiments}
    \label{tab:datasets-models}
    \begin{tabular}{ccccc}
        \hline
        Dataset           & Task                      & Model     & \# of Classes & License      \\ \hline
        Synthetic dataset & Regression                & Linear    & N/A           & N/A          \\
        CIFAR-10          & Image classification      & ResNet-18 & 10            & MIT License  \\
        CIFAR-100         & Image classification      & ResNet-18 & 100           & MIT License  \\
        FEMNIST           & Image classification      & CNN       & 62            & BSD 2-Clause \\
        Shakespeare       & Next character prediction & LSTM      & 79            & BSD 2-Clause \\
        \hline
    \end{tabular}
\end{table}
\begin{table}
    \caption{Architecture of LSTM}
    \label{tab:lstm_architecture}
    \centering
    \begin{tabular}{ccc}
        \toprule
        Layer     & Output Shape & \# of Params \\
        \midrule
        Input     & [80]         & 0            \\
        Embedding & [80, 256]    & 20,224       \\
        LSTM      & [80, 256]    & 1,052,672    \\
        Dropout   & [80, 256]    & 0            \\
        Dense     & [256, 79]    & 20,303       \\
        \bottomrule
    \end{tabular}
\end{table}
\begin{table}
    \caption{Architecture of CNN}
    \label{tab:cnn_architecture}
    \centering
    \begin{tabular}{cccc}
        \toprule
        Layer   & Output Shape & \# of Params & Kernel Size \\
        \midrule
        Input   & [1, 28, 28]  & 0            &             \\
        Conv2d  & [32, 26, 26] & 320          & (3, 3)      \\
        Conv2d  & [64, 24, 24] & 18,496       & (3, 3)      \\
        Dropout & [64, 24, 24] & 0            &             \\
        Dense   & [128]        & 1,179,776    &             \\
        Dropout & [128]        & 0            &             \\
        Dense   & [62]         & 7,998        &             \\
        \bottomrule
    \end{tabular}
\end{table}

\paragraph{Synthetic dataset}
Here, we briefly describe the synthetic dataset used in our experiments.
For client $i=1, \dots, 20$, we generate 30 samples $\{(x_j, y_j)\}~(x_j \in \R^{1000})$
by sampling $x_j \sim \mathcal{N}(0, \Sigma)$ and $y_j = \langle w_{i, j}, x_j \rangle$.
Here, $\Sigma$ is a diagonal matrix with its $k$-th diagonal element $\Sigma_{k, k} = k^{-1.1}$,
and $w_{i, j} \sim \mathcal{N}(w_i, 1), w_i \sim \mathcal{N}(0, 0.1)$.

\subsection{Hyperparameters}
For a fair comparison, we tune the hyperparameters for each method using grid search.
We run algorithms for 500 rounds for the synthetic dataset and 50 rounds for the real-world datasets, and employ the hyperparameters which yield the best validation accuracy averaged over the last 5 rounds.
We summarize the best hyperparameters in Table~\ref{table:hyperparameters}.
Other hyperparameters are kept the same across all methods.
Following~\citet{jhunjhunwala2023fedexp}, we use weight decay of $10^{-4}$, learning rate decay of $0.998$, and gradient clipping to stabilize the training for image classification tasks.
\begin{table}
    \caption{Best hyperparameters (log$_{10}$ scale)}
    \label{table:hyperparameters}
    \centering
    \begin{tabular}{ccccccccc}
        \toprule
                    & \multicolumn{2}{c}{FedAvg} & \multicolumn{2}{c}{FedExp} & \multicolumn{2}{c}{FedAvgM} & \multicolumn{2}{c}{FedExpM}                                                 \\
        dataset     & $\eta_l$                   & $\eta_g$                   & $\eta_l$                    & $\epsilon_g$                & $\eta_l$ & $\eta_g$ & $\eta_l$ & $\epsilon_g$ \\
        \midrule
        CIFAR100    & -2                         & 0                          & -2                          & -4                          & -2       & 0        & -2       & -3           \\
        CIFAR10     & -2                         & 0                          & -2                          & -4                          & -2       & -1       & -2       & -4           \\
        FEMNIST     & -1                         & 0                          & -1                          & -2                          & -2       & 0        & -1       & -3           \\
        shakespeare & 0                          & 0                          & 0                           & -2                          & 0        & 0        & 0        & -4           \\
        \bottomrule
    \end{tabular}
    \begin{tabular}{ccccccccc}
        \toprule
                    & \multicolumn{2}{c}{FedAdagrad} & \multicolumn{2}{c}{FedDuAdagrad} & \multicolumn{2}{c}{FedAdam} & \multicolumn{2}{c}{FedDuAdam}                                                 \\
        dataset     & $\eta_l$                       & $\eta_g$                         & $\eta_l$                    & $\epsilon_g$                  & $\eta_l$ & $\eta_g$ & $\eta_l$ & $\epsilon_g$ \\
        \midrule
        CIFAR100    & -2                             & -4                               & -2                          & -1                            & -2       & -4       & -2       & -1           \\
        CIFAR10     & -2                             & -4                               & -2                          & -1                            & -2       & -4       & -2       & -2           \\
        FEMNIST     & -1                             & -2                               & -1                          & -1                            & -2       & -2       & -1       & -1           \\
        shakespeare & 0                              & -2                               & 0                           & 0                             & 0        & -2       & 0        & 0            \\
        \bottomrule
    \end{tabular}
\end{table}

\paragraph{Synthetic dataset}
For the synthetic dataset, We tune $\eta_l$ over $\{10^{-3}, 10^{-5/2}, 10^{-2}, 10^{-3/2}, 10^{-1}\}$,
and $\eta_g$ over $\{10^{-1}, 10^{-1/2}, 10^{0}, 10^{1/2}, 10^1\}$ for FedAvg(M) and $\{10^{-2}, 10^{-3/2}, 10^{-1}, 10^{-1/2}, 10^{0}\}$ for FedOPT.
We fix $\epsilon=0, \epsilon_g=0$.

\paragraph{Image classification datasets}
For the image classification tasks, we tune $\eta_l$ over $\{10^{-2}, 10^{-3/2}, 10^{-1}, 10^{-1/2}, 10^{0}\}$.
The grid of $\eta_g$ is $\{10^{-1}, 10^{-1/2}, 10^0, 10^{1/2}, 10^1\}$ for FedAvg(M) and SCAFFOLD,
and $\{10^{-4}, 10^{-7/2}, 10^{-3}, 10^{-5/2}, 10^{-2}\}$ for FedOPT.
The grid of $\epsilon_g$ is $\{10^{-3}, 10^{-5/2}, 10^{-2}, 10^{-3/2}, 10^{-1}\}$ for FedDuA,
and $\{10^{-4}, 10^{-7/2}, 10^{-3}, 10^{-5/2}, 10^{-2}\}$ for FedExP(M).
We fix $\epsilon=10^{-9}$ for adaptive methods if not specified.

\paragraph{NLP dataset}
For the NLP task, we tune $\eta_l$ over $\{10^{-2}, 10^{-3/2}, 10^{-1}, 10^{-1/2}, 10^{0}\}$.
The grid of $\eta_g$ is $\{10^{-1}, 10^{-1/2}, 10^0, 10^{1/2}, 10^1\}$ for FedAvg(M) and SCAFFOLD,
and $\{10^{-3}, 10^{-5/2}, 10^{-2}, 10^{-3/2}, 10^{-1}\}$ for FedOPT.
The grid of $\epsilon_g$ is $\{10^{-3}, 10^{-5/2}, 10^{-2}, 10^{-3/2}, 10^{-1}\}$ for FedDuA,
and $\{10^{-1}, 10^{-1/2}, 10^{0}, 10^{1/2}, 10^{1}\}$ for FedExP(M).
We fix $\epsilon=10^{-9}$ for adaptive methods if not specified.

\section{Additional Experimental Results}\label{appendix:additional-experimental-results}
\subsection{Long-term Behavior}\label{appendix:long-term}
\begin{figure}[t]
    \centering
    \includegraphics[width=0.95\textwidth]{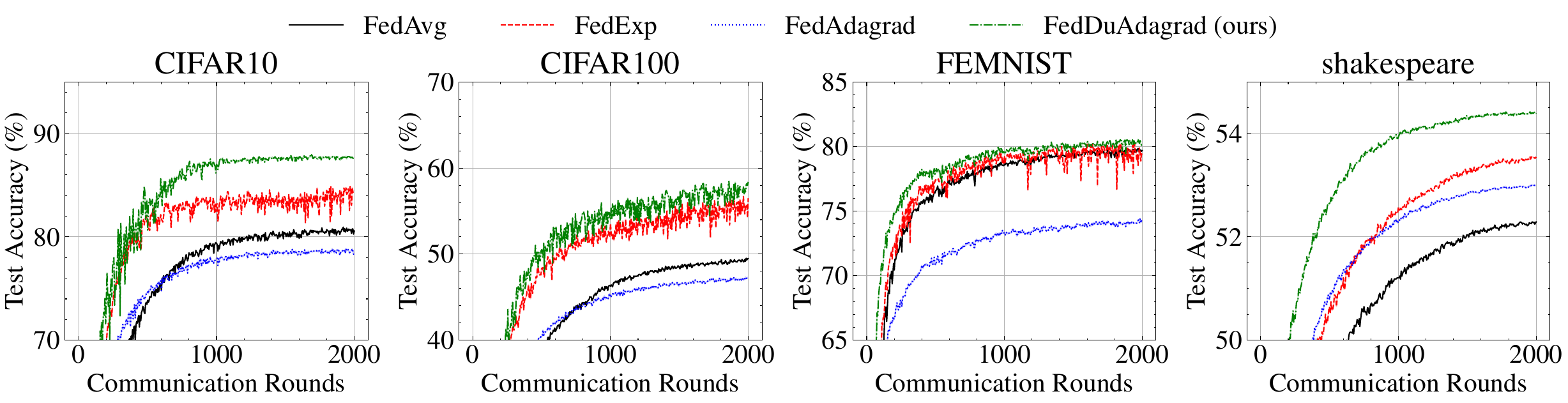}
    \includegraphics[width=0.95\textwidth]{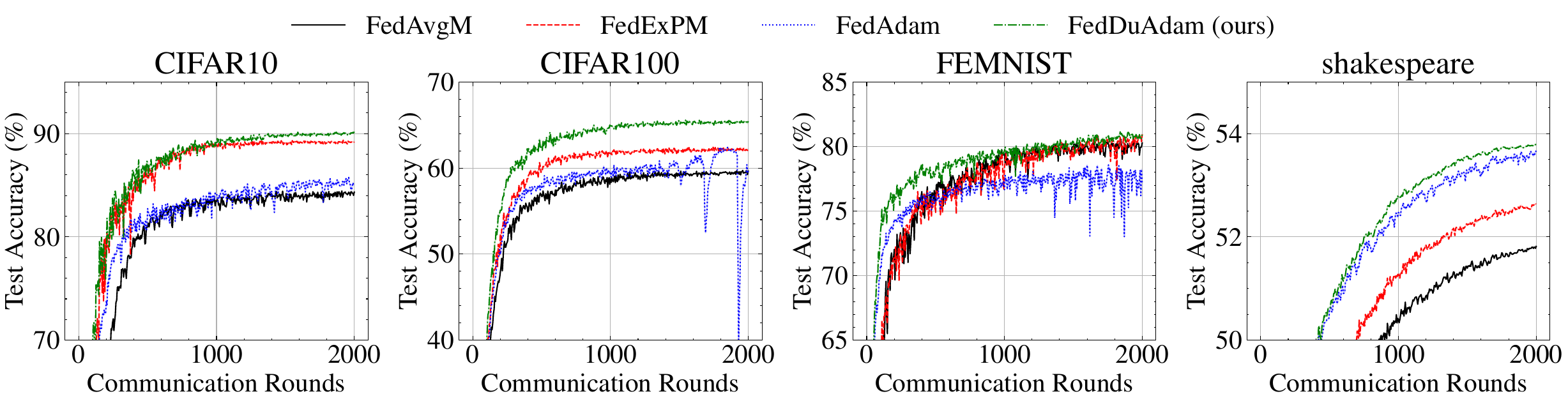}
    \caption{Long term behavior of each algorithm.}
    \label{fig:long-term}
\end{figure}

To compare the long-term behavior of our proposed method and baselines, we ran the experiments for 2000 rounds, which is sufficiently long to observe the convergence behavior.
We show the results in Fig.~\ref{fig:long-term}.
We see that $\fedada$ consistently outperforms other methods in terms of convergence speed and final accuracy.

\subsection{Comparison with FedProx}\label{appendix:prox}
\begin{wrapfigure}{r}[0pt]{0.25\textwidth}
    \vspace{-3em}
    \centering
    \includegraphics[width=0.24\textwidth]{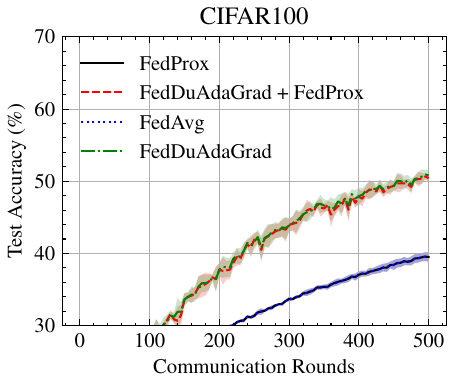}
    \caption{Comparison with FedProx.}
    \label{fig:prox}
    \vspace{-3em}
\end{wrapfigure}
In this section, we provide a comparison with FedProx~\citep{li2020federated}.
This is an algorithm that modifies the local objective and thus can be combined with the $\fedada$ framework.
For the FedProx-type local training procedure, we tune the additional hyperparameter $\mu$ with the grid $\{10^{-3}, 10^{-2}, 10^{-1}, 1\}$
following the original paper.
As shown in Fig.~\ref{fig:prox}, FedProx-type local training does not improve performance in our setup.

\vspace{1.5em}
\subsection{Training Loss and Global Learning Rate}\label{appendix:training-loss}
\begin{figure}[h]
    \centering
    \includegraphics[width=0.95\textwidth]{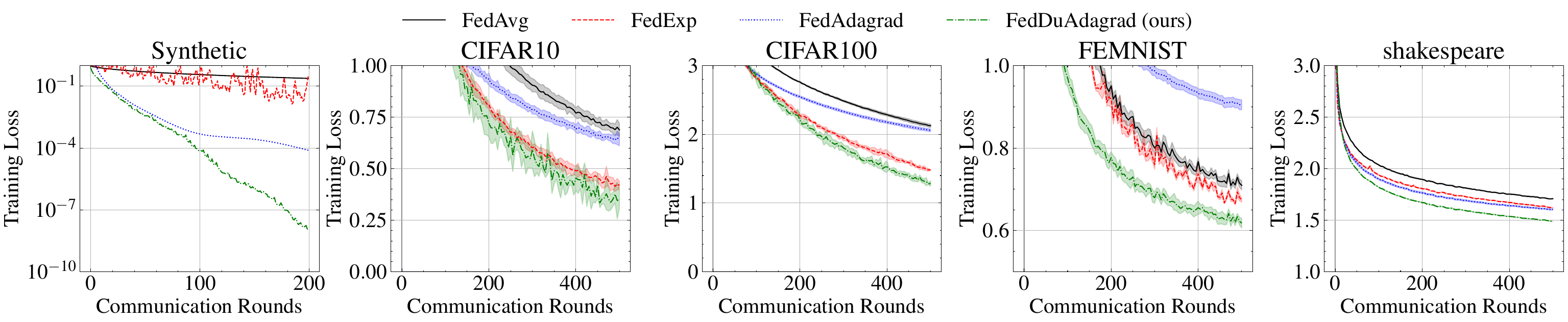}
    \includegraphics[width=0.95\textwidth]{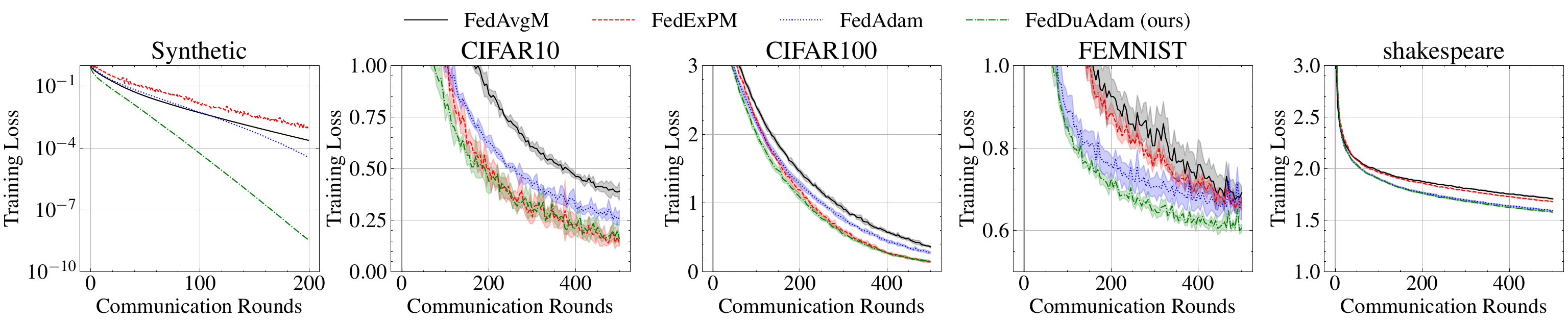}
    \caption{Comparison of training loss for different methods.}
    \label{fig:training-loss}
\end{figure}

Here, we provide the curve of training loss and global learning rate for $\fedada$ and baselines.
As shown in Fig.~\ref{fig:training-loss}, $\fedada$ converges faster than other methods in terms of training loss.
\begin{figure}[h]
    \centering
    \includegraphics[width=0.95\textwidth]{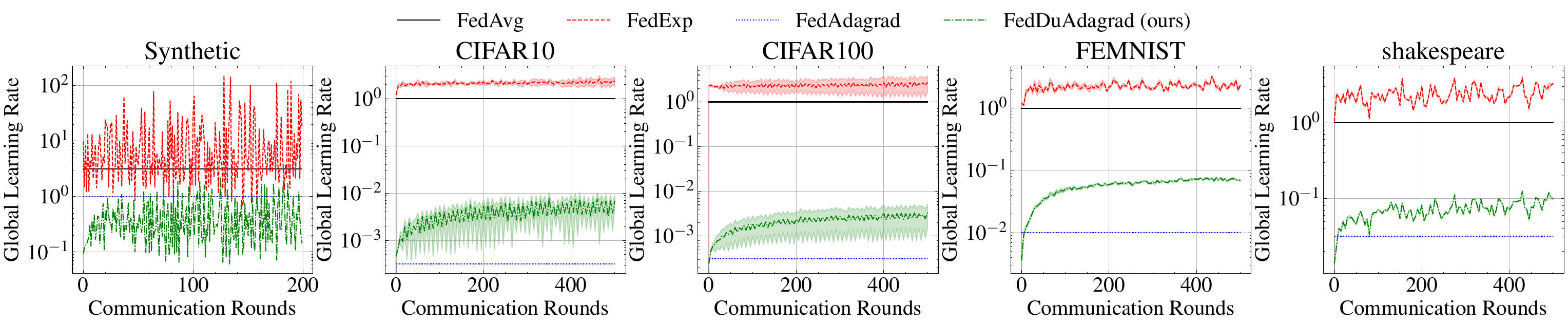}
    \includegraphics[width=0.95\textwidth]{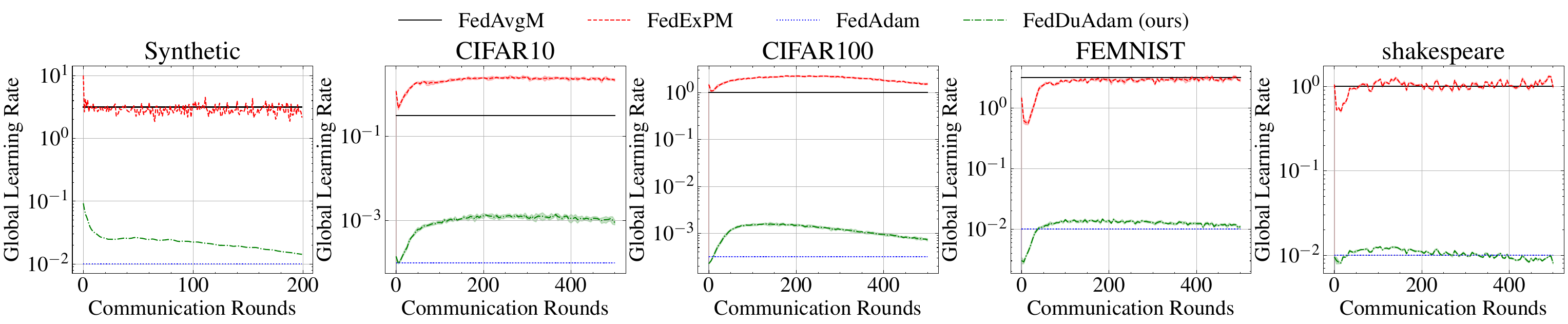}
    \caption{Comparison of global learning rates $\eta_g$ for different methods.}
    \label{fig:global-learning-rate}
\end{figure}

\section{Limitations and Broader Impacts}\label{appendix:limitations-broader-impacts}
\paragraph{Limitations}
Although our proposed method is applicable and numerically shown to be effective in client subsampling setting,
theoretical analysis is difficult since the global learning rate is stochastic in this case.
This has been discussed in some previous works~\citep{jhunjhunwala2023fedexp,li2024the}
and not limited to our work.
In addition, our theoretical analysis focuses on the case of convex objectives
while objective functions are often non-convex in deep learning.
Though we empirically show that our method works well in non-convex settings,
the theoretical analysis is still an open problem.
We leave extending our theoretical analysis to the above cases as future work.
\paragraph{Broader Impacts}
This work can be applied to various real-world applications, which is beneficial for the society.
For instance, by enabling more efficient and robust federated learning,
this work could contribute to advancements in privacy-preserving collaborative AI model development
in sensitive domains like healthcare, or enhance the personalization of services on edge devices while protecting user data.
On the other hand, our proposed method can be used for training malicious models.
While this work focuses on algorithmic advancements,
the authors acknowledge the ongoing community efforts towards responsible AI development and encourage the use of such technologies in an ethical and beneficial manner.
\end{document}